\documentclass[11pt,a4paper]{article}
\usepackage[hyperref]{emnlp-ijcnlp-2019}
\usepackage{times}
\usepackage{latexsym}

\usepackage{url}
\usepackage{drs}
\usepackage{arydshln}

\usepackage{amsfonts}
\usepackage{amsmath}
\usepackage{amssymb}

\aclfinalcopy 

\title{A survey of cross-lingual features for \\ zero-shot cross-lingual semantic parsing}

\author{Jingfeng Yang$^{\ddagger}$* \quad Federico Fancellu$^{\dagger}$* \quad Bonnie Webber$^{\diamond}$\\
  $^{\ddagger}$ Georgia Institute of Technology\\
  $^\dagger$ Samsung AI Center, Toronto\\
  $^\diamond$ ILCC, School of Informatics, University of Edinburgh\\
  {\tt jyang690@gatech.edu \quad \tt federico.f@samsung.com \quad \tt bonnie@inf.ed.ac.uk}
  }

\date{}

\begin{document}
\maketitle
\begin{abstract}
The availability of corpora to train semantic parsers in English has lead to significant advances in the field. Unfortunately, for languages other than English, annotation is scarce and so are developed parsers. We then ask: could a parser trained in English be applied to language that it hasn't been trained on? To answer this question we explore zero-shot cross-lingual semantic parsing where we train an available coarse-to-fine semantic parser \citep{liu2018discourse} using cross-lingual word embeddings and universal dependencies in English and test it on Italian, German and Dutch. Results on the Parallel Meaning Bank -- a multilingual semantic graphbank, show that Universal Dependency features significantly boost performance when used in conjunction with other lexical features but modeling the UD structure directly when encoding the input does not.
\end{abstract}

\section{Introduction}

Semantic parsing is a task of transducing natural language to meaning representations, which in turn can be expressed through many different semantic formalisms including lambda calculus \cite{zettlemoyer2012learning}, DCS \cite{liang2013learning}, Discourse Representation Theory (DRT) \cite{kamp2013discourse}, AMR \cite{banarescu2013abstract} and so on. This availability of annotated data in English has translated into the development of a plethora of models, including encoder-decoders \cite{dong2016language,DBLP:conf/acl/JiaL16} as well as tree or graph-structured decoders \cite{dong2016language,dong2018coarse,liu2018discourse,yin2017syntactic}. {\let\thefootnote\relax\footnote{{*Work done when Jingfeng Yang was an intern and Federico Fancellu a post-doc at the University of Edinburgh}}}

Whereas the majority of semantic banks focus on English, recent effort has focussed on building multilingual representations, e.g. PMB \cite{abzianidze2017parallel}, MRS \cite{copestake1995translation} and FrameNet\cite{pado2005cross}. However, manually annotating meaning representations in a new language is a painstaking process which explains why there are only a few datasets available for different formalisms in languages other than English. As a consequence, whereas the field has made great advances for English, little work has been done in other languages.

\begin{figure}
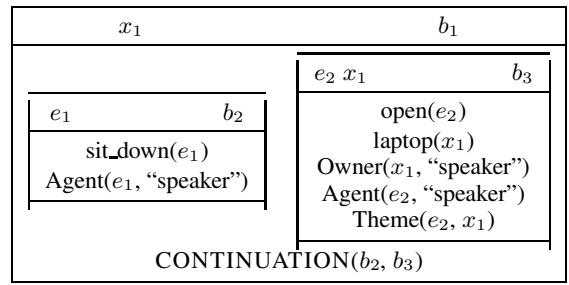
\footnotesize
\drs{$x_1$ \qquad\qquad\qquad\qquad\qquad\qquad $b_1$}{\condrs{$e_1$ \qquad\qquad\qquad $b_2$}{sit\_down($e_1$)\\Agent($e_1$, ``speaker'')}{}{$e_2$ $x_1$\quad\quad\quad\quad\quad\quad $b_3$}{open($e_2$)\\laptop($x_1$)\\Owner($x_1$, ``speaker'')\\Agent($e_2$, ``speaker'')\\Theme($e_2$, $x_1$)}\\\textsc{CONTINUATION($b_2$, $b_3$)}}
\caption{The Discourse Representation Structure (DRS) for ``\textsl{I sat down and opened my laptop}''. For simplicity, we have omitted any time reference.}\label{fig:drs}
\end{figure}

We ask: \textit{can we learn a semantic parser for English and test it where in another where annotations are not available? What would that require?}

To answer this question, previous work have leveraged machine translation techniques to map the semantics from a language to another \citep[e.g.][]{damonte2018cross}. However, these methods require parallel corpora to extract automatic alignments which are often noisy or not available at all.

In this paper we explore parameter-shared models instead, where a model is trained on English using language independent features and tested in a target language.

To show how this approach performs, we focus on the Parallel Meaning Bank \cite[PMB][]{abzianidze2017parallel} -- a multilingual semantic bank, where sentences in English, German, Italian and Dutch have been annotated with their meaning representations. The annotations in the PMB are based on Discourse Representation Theory \citep[DRT,][]{kamp2013discourse}, a popular theory of meaning representation designed to account for intra and inter-sentential phenomena, like temporal expressions and anaphora. Figure 1 shows an example DRT for the sentence `I sat down and opened my laptop' in its canonical `box' representation. A DRS is a nested structure with the top part containing the discourse references and the bottom with unary and binary predicates, as well as semantic constants (e.g. `speaker'). DRS can be linked to each other via logic operator (e.g. $\lnot$, $\rightarrow$, $\diamond$) or, as in this case, discourse relations (e.g. CONTINUATION, RESULT, ELABORATION, etc.).

To test our approach we leverage the DRT parser of \newcite{liu2018discourse}, an encoder-decoder architecture where the meaning representation is reconstructed in three stages, coarse-to-fine, by first building the DRS skeleton (i.e. the `box' structures) and then fill each DRS with predicates and variables. Whereas the original parser utilizes a sequential Bi-LSTM encoder with monolingual lexical features, we experiment with language-independent features in the form of cross-lingual word-embeddings, universal PoS tags and universal dependencies. In particular, we also make use of tree encoders to assess whether modelling syntax can be beneficial in cross-lingual settings, as shown for other semantic tasks (e.g. negation scope detection \cite{fancellu2018neural}).

Results show that language-independent features are a valid alternative to projection methods for cross-lingual semantic parsing. We show that adding dependency relation as features is beneficial, even when they are the only feature used during encoding. However, we also show that modeling the dependency structure directly via tree encoders does not outperform a sequential BiLSTM architecture for the three languages we have experimented with.

\section{Methods}
\subsection{Model}
In this section, we describe the modifications to the coarse-to-fine encoder-decoder architecture of \citet{liu2018discourse}; for more detail, we refer the reader to the original paper.

\subsubsection{Encoder}

\textbf{BiLSTM}. We use \citet{liu2018discourse}'s Bi-LSTM as baseline. However, whereas the original model represents each token in the input sentence as the concatenation of word ($e_{w_i}$) and lemma embeddings, we discard the latter and add a POS tag embedding ($e_{p_i}$) and dependency relation embedding ($e_{d_i}$) feature. These embeddings are concatenated to represent the input token. The final encoder representation is obtained by concatenating both final forward and backward hidden states.

\noindent\textbf{TreeLSTM}. To model the dependency structure directly, we use a child-sum tree-LSTM \cite{tai2015improved}, where each word in the input sentence corresponds to a node in the dependency tree. In particular, summing across children is advantageous for cross-lingual tasks since languages might display different word orders. Computation follows Equation (1).
\begin{equation}
\begin{split}
&x_i=tanh([e_{w_i};e_{p_i};e_{d_i}]W_1+b_1)\\
&[h_{e_1}:h_{e_n}]=TreeLSTM(x_1:x_n)
\end{split}
\end{equation}

\noindent\textbf{Po/treeLSTM}. Completely discarding word order might hurt performance for related languages, where a soft notion of positioning can help. To this end, we add a positional embeddings $P_i$ \cite{vaswani2017attention} that helps the child-sum tree-LSTM discriminating between the left and right child of a parent node. This is computed following Equation (2) where $i$ is the position of the word, $j$ is the $jth$ dimension in total $d$ dimensions.
\begin{equation}
\begin{split}
&P_{(i,2j)}=sin(i/1000^{2j/d})\\
&P_{(i,2j+1)}=cos(i/1000^{2j/d})\\
&x_i=tanh([e_{w_i}+P_i ; e_{p_i}+P_i ; e_{d_i}+P_i]W_1+b_1)\\
&[h_{e_1}:h_{e_n}]=TreeLSTM(x_1:x_n)
\end{split}
\end{equation}
\textbf{Bi/treeLSTM}. Finally, similarly to \newcite{chen2017improved}, we combine tree-LSTM and Bi-LSTM, where a tree-LSTM come is initialized using the last layer of a Bi-LSTM, which encodes order information. Computation is shown in Equation (3).
 \begin{equation}
 \begin{split}
 &x_i=tanh([e_{w_i};e_{p_i}]W_1+b_1)\\
 &[\bar{h}_1:\bar{h}_n]=BiLSTM(x_1:x_n)\\
 &[h_{e_1}:h_{e_n}]=TreeLSTM([\bar{h}_1;e_{d_1}]:[\bar{h}_n;e_{d_n}])
 \end{split}
 \end{equation}

\subsubsection{Decoder}
The decoder of \newcite{liu2018discourse} reconstructs the DRS in three steps, by first predicting the overall structure (the `boxes'), then the predicates and finally the referents, with each subsequent step being conditioned on the output of the previous. During predicate prediction, the decoder uses a copying mechanism to predict those unary predicates that are also lemmas in the input sentence (e.g. `eat'). For the those that are not, soft attention is used instead. No modifications were done to the decoder; for more detail, we refer the reader to the original paper.

\subsection{Data}
We use the PMB v.2.1.0 for the experiments. The dataset consists of 4405 English sentences, 1173 German sentences, 633 Italian sentences and 583 Dutch sentences. We divide the English sentences into 3072 training sentences, 663 development and 670 testing sentences. We consider all the sentences in other languages as test set.

In order to be used as input to the parser, \newcite{liu2018discourse} first convert the DRS into tree-based representations, which are subsequently linearized into PTB-style bracketed sequences. This transformation is lossless in that re-entrancies are duplicated to fit in the tree structure. We use the same conversion in this work; for further detail we refer the reader to the original paper.

Finally, it is worth noting that lexical predicates in PMB are in English, even for non-English languages. Since this is not compatible with our copy mechanism, we revert predicates to their original language by substituting them with the lemmas of the tokens they are aligned to (since gold alignment information is included in the PMB).

\subsection{Cross-lingual features}
In order to make the model directly transferable to the German, Italian and Dutch test data, we use the following language-independent features.

\textbf{Multilingual word embeddings}. We use the MUSE \cite{conneau2017word} pre-trained multilingual word embeddings and keep them fixed during training. 

\textbf{UD relations and structure}. We use UDPipe \cite{udpipe:2017} to obtain parses for English, German, Italian and Dutch. UD relation embeddings are randomly initialized and updated.

\textbf{Universal POS tags}. We use the Universal POS tags \cite{petrov2011universal} obtained with UDPipe parser. Universal POS tag embeddings are randomly initialized and updated during training.

\subsection{Model comparison}
We use the BiLSTM model as baseline (\textit{Bi}) and compare it to the child-sum tree-LSTM (\textit{tree}) with positional information added (\textit{Po/tree}), as well as to a treeLSTM initialized with the hidden states of the BiLSTM(\textit{Bi/tree}). We also conduct an ablation study on the features used, where \textit{WE}, \textit{PE} and \textit{DE} are the word-embedding, PoS embedding and dependency relation embedding respectively. For completeness, along with the results for the cross-lingual task, we also report results for monolingual English semantic parsing, where word embedding features are randomly initialized.

\subsection{Evaluation}
We use Counter \cite{van2018evaluating} to evaluate the performance of our models. Counter looks for the best alignment between the predicted and gold DRS and computes precision, recall and F1. For further details about Counter, the reader is referred to \newcite{van2018evaluating}. It is worth reminding that unlike other work on the PMB \citep[e.g.][]{van2018exploring}, \citet{liu2018discourse} does not deal with presupposition. In the PMB, presupposed variables are extracted from a main box and included in a separate one. In our work, we revert this process so to ignore presupposed boxes. Similarly, we also do not deal with sense tags which we aim to include in future work.

\section{Results and Analysis}
\begin{table*}
\footnotesize
\centering
\begin{tabular}{c|c|c|c|c|c|c|c|c|c}
Model  & \multicolumn{3}{c|}{German} & \multicolumn{3}{c|}{Italian} & \multicolumn{3}{c}{Dutch} \\\hline
&P&R&F&P&R&F&P&R&F\\\hdashline
$Bi_{WE,PE}$ & 0.4996& \textbf{0.4614}& 0.4797 & 0.5102& \textbf{0.5319} & 0.5208& 0.4219& 0.4780& 0.4482 \\
$tree_{WE,PE}$  & 0.4457& 0.375& 0.4075 & 0.5088& 0.4257 & 0.4636& 0.4627& 0.3592& 0.4044\\
 $Po/tree_{WE,PE}$   & \textbf{0.5911}& 0.4546& \textbf{0.5139} & \textbf{0.5955}& 0.4894&\textbf{ 0.5373} & \textbf{0.5027} & 0.4296& 0.4633 \\
$Bi/tree_{WE,PE}$  & 0.5482& 0.4587& 0.4995 &0.4986&0.5498&0.5229 & 0.4627& \textbf{0.4943}&\textbf{ 0.4780}\\\hdashline

$Bi_{WE,PE,DE}$  & 0.6763 & 0.6060& 0.6392& \textbf{0.7129}&\textbf{0.6669}&\textbf{0.6891} & \textbf{0.6286}& 0.5381 & \textbf{0.5798}\\
$tree_{WE,PE,DE}$  &\textbf{0.6767}& \textbf{0.6080}& \textbf{0.6405} & 0.6885& 0.6429& 0.6649 & 0.5926& \textbf{0.5437} & 0.5690\\
$Po/tree_{WE,PE,DE}$ & 0.6750 & 0.5280& 0.5925& 0.6724& 0.5637 & 0.6133& 0.6096 & 0.4728& 0.5360\\
$Bi/tree_{WE,PE,DE}$  & 0.6496& 0.5950 & 0.6211&0.6534 &0.6393&0.6463& 0.5722& 0.5369& 0.5540\\\hdashline
$Bi_{DE}$ & 0.6532& \textbf{0.6290}& \textbf{0.6409} & 0.6926& \textbf{0.6749} & \textbf{0.6836}& 0.5792& 0.5318& 0.5545 \\
$tree_{DE}$ &\textbf{0.6695}&0.5822&0.6228&\textbf{0.6965}&0.6133&0.6523 & \textbf{0.6048}& 0.5609 & \textbf{0.5820}\\
 $Po/tree_{DE}$  & 0.6453& 0.6250& 0.6350 & 0.6896&0.6622& 0.6756 & 0.5915 & \textbf{0.5671}& 0.5790 \\
$Bi/tree_{DE}$  &  --&  --& --& --& --& --& --& --& --\\\hdashline
$Bi_{WE,DE}$& 0.6708& 0.5921& 0.6290 & 0.6997& \textbf{0.7002} & \textbf{0.6999}& 0.6202& \textbf{0.5919}& \textbf{0.6057} \\
$tree_{WE,DE}$ &0.6466&\textbf{0.6335}&0.6400&0.7072&0.6902&0.6986 & 0.6070& 0.5729 & 0.5895\\
$Po/tree_{WE,DE}$  & 0.6520& 0.6294& 0.6405 & 0.7079& 0.6793& 0.6933 & \textbf{0.6209} & 0.5828& 0.6012 \\
$Bi/tree_{WE,DE}$ & \textbf{0.6750}& 0.6169& \textbf{0.6446} & \textbf{0.7110}& 0.6622& 0.6857 & 0.6175 & 0.5481& 0.5807 

\end{tabular}
\caption{\label{tab:widgets5}Results of zero-shot cross-lingual semantic parsing for models trained in English and tested in German, Italian and Dutch.\footnote{Given that we need either word or PoS tag to initialize the BiLSTM, a \textit{Bi/tree} model cannot be used when testing with dependency features only (\textit{DE})}}
\end{table*}

\begin{table}[h]
\centering
\begin{tabular}{c|c|c|c}
Model  & P & R & F \\\hline
$Bi_{WE,PE}$  & \textbf{0.8825} &\textbf{0.8453} &\textbf{0.8635} \\
$tree_{WE,PE}$ & 0.8512 &0.8154 &0.8329 \\
$Po_{WE,PE}$  & 0.8592&0.8296&0.8441 \\
$Bi/tree_{WE,PE}$ & 0.8670&0.8433&0.8550\\\hdashline
$Bi_{WE,PE,DE}$  & \textbf{0.8919} &\textbf{0.8584} &\textbf{0.8748} \\
$tree_{WE,PE,DE}$  & 0.8590 &0.8362 &0.8474 \\
$Po/tree_{WE,PE,DE}$   & 0.8503&0.8305&0.8403 \\
$Bi/tree_{WE,PE,DE}$ & 0.8602&0.8369&0.8484\\\hdashline
$Bi_{DE}$  & \textbf{0.6629} &0.6417 &0.6521 \\
$tree_{DE}$ & 0.6550 &0.6589 &\textbf{0.6569} \\
$Po/tree_{DE}$ & 0.6522&\textbf{0.6591}&0.6556 \\
$Bi/tree_{DE}$ & --&--&--\\\hdashline
$Bi_{WE,DE}$  & \textbf{0.8764} &\textbf{0.8593} &\textbf{0.8678} \\
$tree_{WE,DE}$ & 0.8569 &0.8356 &0.8461 \\
$Po/tree_{WE,DE}$  & 0.8540&0.8396&0.8467 \\
$Bi/tree_{WE,DE}$ & 0.8655&0.8369&0.8510
\end{tabular}
\caption{\label{tab:widgets1}Results for monolingual semantic parsing (i.e. trained and tested in English)}
\end{table}

\begin{table*}
\small
\centering
\begin{tabular}{c|c|c|c|c|c|c|c|c|c}
 & \multicolumn{3}{c|}{German} & \multicolumn{3}{c|}{Italian} & \multicolumn{3}{c}{Dutch} \\\hline
&P&R&F&P&R&F&P&R&F\\\hline
operators &  0.7158&  0.3778 &  0.4945& 0.9302& 0.3846& 0.5442 & 0.5833 & 0.1892& 0.2857\\\hdashline
non-lexical predicate & 0.6507& 0.5887 & 0.6182 &0.6625&0.6848&0.6735 & 0.6468& 0.5970& 0.6209 \\
\emph{\small{unary}} & 0.7700& 0.6641& 0.7131 &0.7974&0.7730&0.7850& 0.7615& 0.6645& 0.7097\\
\emph{\small{binary}} & 0.5626& 0.5281& 0.5448 &0.5627&0.6117&0.5862& 0.5640& 0.5433& 0.5535\\\hdashline
lexical predicate &  0.7286&  0.7326&  0.7306 & 0.6622& 0.7705& 0.7123 & 0.4833& 0.6070& 0.5381
\end{tabular}
\caption{\label{tab:widgets6}Error analysis.}
\end{table*}

Table~\ref{tab:widgets5} shows the performance of our cross-lingual models in German, Italian and Dutch. We summarize the results as follows:

\textbf{Dependency features are crucial for zero-shot cross-lingual semantic parsing}. Adding dependency features dramatically improves the performance in all three languages, when compared to using multilingual word-embedding and universal PoS embeddings alone. We hypothesize that the quality of the multilingual word-embeddings is poor, given that models using embeddings for the dependency relations alone outperform those using the other two features.

\textbf{TreeLSTMs slightly improve performance only for German}. TreeLSTMs do not outperform a baseline BiLSTM for Italian and Dutch and they show little improvement in performance for German. This might be due to different factors that deserve more analysis including the performance of the parsers and syntactic similarity between these languages. When only dependency features are available, we found treeLSTM to boost performance only for Dutch.

\textbf{BiLSTM are still state-of-the-art for monolingual semantic parsing for English}. Table \ref{tab:widgets1} shows the result for the models trained and tested in English. Dependency features in conjunction with word and PoS embeddings lead to the best performance; however, in all settings explored treeLSTMs do not outperform a BiLSTM.

 \subsection{Error Analysis}
We perform an error analysis to assess the quality of the prediction for \textit{operators} (i.e. logic operators like ``Not'' as well as discourse relations ``Contrast''), non-lexical predicates, such as binary predicates (e.g. Agent(e,x)) as well as unary predicates (e.g. time(t), entity(x), etc.), as well as for lexical predicates (e.g. open(e)). Results in Table ~\ref{tab:widgets6} show that predicting operators and binary predicates across language is hard, compared to the other two categories. Prediction of lexical predicates is relatively good even though most tokens in the test set where never seen during training; this can be attributable to the copy mechanism that is able to transfer tokens from the input directly during predication.

\section{Related work}
Previous work have explored two main methods for cross-lingual semantic parsing. One method requires parallel corpora to extract alignments between source and  target languages using machine translation \cite{pado2005cross,damonte2017cross,zhang2018cross}
The other method is to use parameter-shared models in the target language and the source language by leveraging language-independent features such as multilingual word embeddings, Universal POS tags and UD \cite{reddy2017universal,duong2017multilingual,susanto2017neural,mulcaire2018polyglot}. 

For semantic parsing, encoder-decoder models have achieved great success. Amongst these, tree or  graph-structured  decoders  have recently shown to be state-of-the-art \cite{dong2016language,dong2018coarse,liu2018discourse,cheng2017learning,yin2017syntactic}.
\section{Conclusions}
We go back to the questions in the introduction:

\textit{Can we train a semantic parser in a language where annotation is available?}. In this paper we show that this is indeed possible and we propose a zero-shot cross-lingual semantic parsing method based on language-independent features, where a parser trained in English -- where labelled data is available, is used to parse sentences in three languages, Italian, German and Dutch.

\textit{What would that require?} We show that universal dependency features can dramatically improve the performance of a cross-lingual semantic parser but modelling the tree structure directly does not outperform sequential BiLSTM architectures, not even when the two are combined together.

We are planning to extend this initial survey to other DRS parsers that does not exclude presupposition and sense as well as to other semantic formalisms (e.g. AMR, MRS) where data sets annotated in languages other than English are available. Finally, we want to understand whether adding a bidirectionality to the treeLSTM will help improving the performance on modelling the dependency structure directly. 

\section*{Acknowledgements}
This work was done while Federico Fancellu was a post-doctoral researcher at the University of Edinburgh. The views expressed are his own and do not necessarily represent the views of Samsung Research.

\bibliography{emnlp-ijcnlp-2019}
\bibliographystyle{acl_natbib}

\end{document}